\DocumentMetadata{}
\documentclass[sigconf,9pt]{acmart}

\AtBeginDocument{%
  \providecommand\BibTeX{{%
    \normalfont B\kern-0.5em{\scshape i\kern-0.25em b}\kern-0.8em\TeX}}}

\copyrightyear{2025}
\acmYear{2025}
\setcopyright{cc}
\setcctype{by}
\acmConference[SenSys '25]{The 23rd ACM Conference on Embedded Networked Sensor Systems}{May 6--9, 2025}{Irvine, CA, USA}
\acmBooktitle{The 23rd ACM Conference on Embedded Networked Sensor Systems (SenSys '25), May 6--9, 2025, Irvine, CA, USA}
\acmDOI{10.1145/3715014.3722081}
\acmISBN{979-8-4007-1479-5/2025/05}

\usepackage[normalem]{ulem} 
\usepackage{xcolor}
\usepackage{color}
\usepackage{graphics}
\usepackage{graphicx}
\usepackage{url}
\usepackage{multirow}
\usepackage{xspace}
\usepackage{enumitem}
\usepackage{subcaption}
\usepackage{balance}
\usepackage{acmart-taps}
\usepackage{algorithm}
\usepackage{algpseudocode}
\usepackage{wasysym}

\usepackage{xspace}
\newcommand{\newname}{FlexiFly\xspace}

\newif\ifedit
\editfalse

\ifedit
    \newcommand{\tp}[1]{
        \vskip 1ex\noindent
        \colorbox{red}{
            \parbox{\columnwidth - 2\fboxsep}{
                \textbf{Issues:} #1
            }
        }
    }
    \newcommand{\preq}[1]{
        \vskip 1ex\noindent
        \colorbox{red}{
            \parbox{\columnwidth - 2\fboxsep}{
                \textbf{Pre-requisites:} #1
            }
        }
    }
    \newcommand{\kaiyuan}[1]{\textcolor{red}{KY: #1}}
    \newcommand{\scott}[1]{\textcolor{red}{MZ: #1}}
    \newcommand{\stephen}[1]{\textcolor{red}{SX: #1}}

    \definecolor{updatecolor}{RGB}{0,102,204} %
    \definecolor{deletecolor}{RGB}{255,0,0}   %
    \definecolor{addcolor}{RGB}{0,153,0}      %
    
    \newcommand{\update}[2]{{\color{updatecolor}\sout{#1} \textbf{#2}}}
    \newcommand{\delete}[1]{{\color{deletecolor}\sout{#1}}}
    \newcommand{\add}[1]{{\color{addcolor}\textbf{#1}}}
    \newcommand{\todo}[1]{\ClassWarning{NOT READY TO SUBMIT}{There is something left todo} \textcolor{red}{[TODO: #1]}}

\else
    \newcommand{\tp}[1]{}    
    \newcommand{\preq}[1]{}    
    \newcommand{\kaiyuan}[1]{}
    \newcommand{\scott}[1]{}
    \newcommand{\stephen}[1]{}

    \newcommand{\update}[2]{#2}
    \newcommand{\delete}[1]{}
    \newcommand{\add}[1]{#1}

    \newcommand{\todo}[1]{\ClassWarning{NOT READY TO SUBMIT}{There is something left todo}}

\fi

\begin{document}
\title[\newname: Interfacing the Physical World with Foundation Models and Reconfigurable Drones]{\newname: Interfacing the Physical World with Foundation Models Empowered by Reconfigurable Drone Systems
}

\author{Minghui Zhao$^*$}
\affiliation{%
  \institution{Columbia University}
  \country{}
}
\email{mz2866@columbia.edu}

\author{Junxi Xia$^*$}
\affiliation{%
  \institution{Northwestern University}
  \country{}
}
\email{junxixia2024@u.northwestern.edu}

\author{Kaiyuan Hou$^*$}
\affiliation{%
  \institution{Columbia University}
  \country{}
}
\email{kh3119@columbia.edu}

\author{Yanchen Liu}
\affiliation{%
  \institution{Columbia University}
  \country{}
}
\email{yl4189@columbia.edu}

\author{Stephen Xia}
\affiliation{%
  \institution{Northwestern University}
  \country{}
}
\email{stephen.xia@northwestern.edu}

\author{Xiaofan Jiang}
\affiliation{%
  \institution{Columbia University}
  \country{}
}
\email{jiang@ee.columbia.edu}

\renewcommand{\shortauthors}{Zhao et al.}

\begin{abstract}
Foundation models (FM) have shown immense human-like capabilities for generating digital media. However, foundation models that can freely sense, interact, and actuate the physical domain is far from being realized. This is due to 1) requiring dense deployments of sensors to fully cover and analyze large spaces, while 2) events often being localized to small areas, making it difficult for FMs to pinpoint relevant areas of interest relevant to the current task. We propose \newname, a platform that enables FMs to ``zoom in'' and analyze relevant areas with higher granularity to better understand the physical environment and carry out tasks. \newname accomplishes by introducing 1) a novel image segmentation technique that aids in identifying relevant locations and 2) a modular and reconfigurable sensing and actuation drone platform that FMs can actuate to ``zoom in'' with relevant sensors and actuators. We demonstrate through real smart home deployments that \newname enables FMs and LLMs to complete diverse tasks up to $85\%$ more successfully. \newname is critical step towards FMs and LLMs that can naturally interface with the physical world.
\end{abstract}

\ccsdesc[500]{Computer systems organization~Embedded and cyber-physical systems}
\ccsdesc[500]{Hardware~Sensors and actuators}
\ccsdesc[500]{Computing methodologies~Scene understanding}
\ccsdesc[500]{Computing methodologies~Cognitive robotics}

\keywords{Embodied AI, Foundation Model Agent, Reconfigurable Drone Platform}

\maketitle
\def\thefootnote{$^*$}\footnotetext{These authors contributed equally to this work}\def\thefootnote{\arabic{footnote}}

\section{Introduction} 
\label{sec:intro_new}

While there are a number of works that incorporate large language models (LLM) into robotic and egocentric systems~\cite{vemprala2023chatgpt,nie2022ai}, such as the Figure01 AGI robot~\cite{figure2024ai} and digital voice assistants, there are few works that explore the use of LLMs and foundation models (FM) to actuate our physical environments. Unlike egocentric systems, where sensing, processing, and control are often localized to the vicinity of the autonomous agent, executing tasks and monitoring spaces often involves sifting through heterogeneous streams of sensing data \textit{spanning large areas} of the space to detect and process events \textit{localized to a tiny fraction}. While FMs and LLMs have shown strong performance on summarization tasks, they have difficulty completing tasks that involve processing localized areas of interest without mechanisms to help them ``zoom in''. Much like how FMs enable general human language and sensory understanding and responses, our work explores how FMs could be used to enable a diverse range of general interactions with objects and spaces that likely cannot be actuated digitally through code. These interactions form a large portion of our daily interactions.

\begin{figure}[t!]
    \centering
    \includegraphics[width=0.9\linewidth]{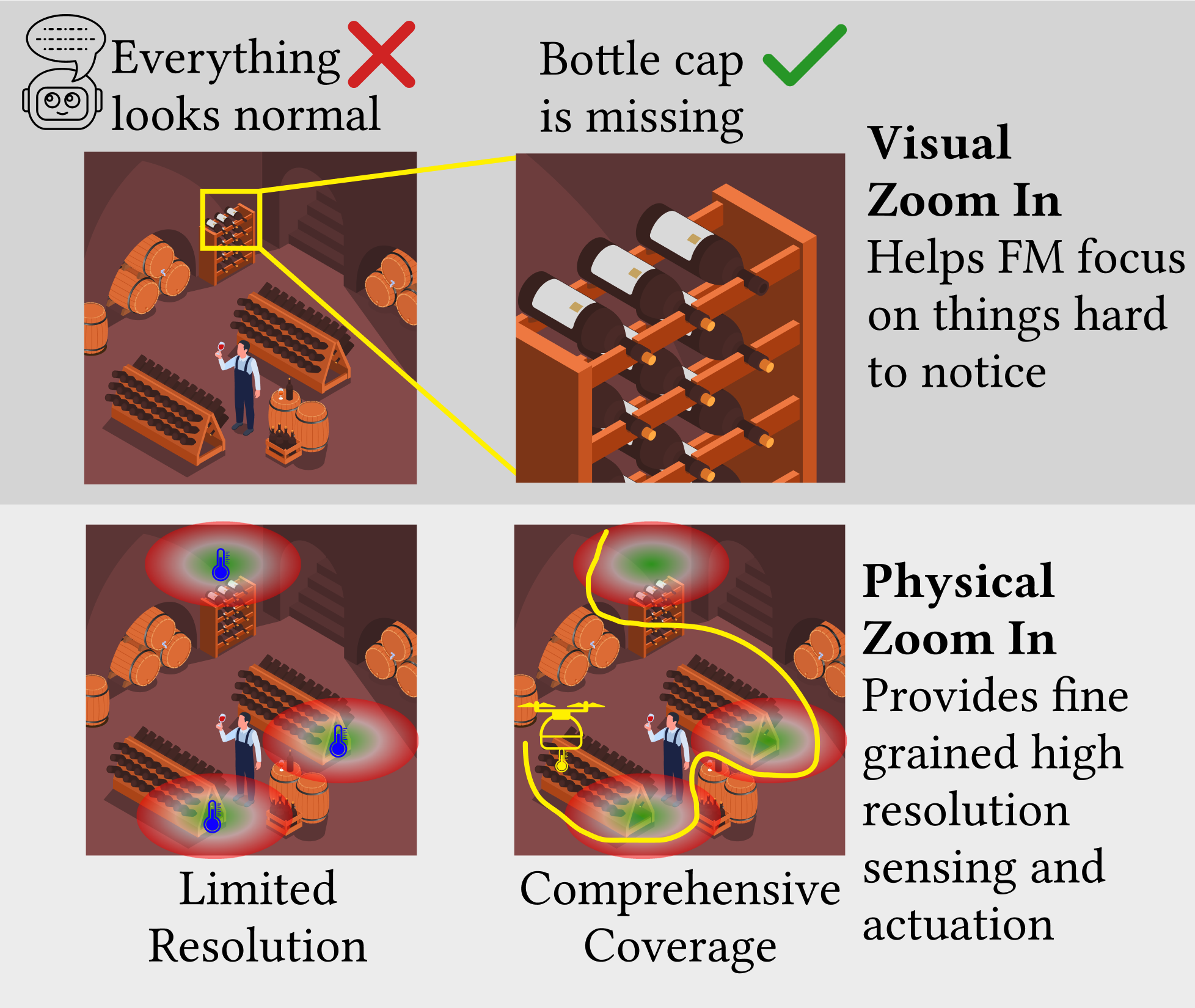}
    \caption{\newname enables FMs to ``zoom in'' to areas of interest with reconfigurable drones to better interface with physical environments.}
    \label{fig:teaser}
\end{figure}

\begin{figure*}[t!]
    \centering
    \includegraphics[width=\linewidth]{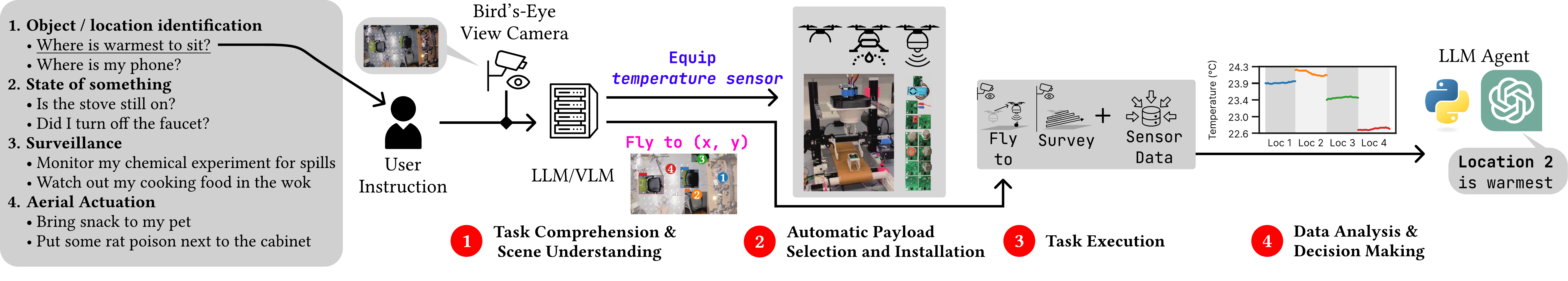}
    \caption{System architecture of intelligent assistant with \newname.}
    \label{fig:personal_assistant_architecture}
\end{figure*}

For example, a person may want his/her home to physically bring a snack or medicine; this would require a robotic system that could physically identify and carry the payload. Another person, coming into an office area for the time may want to know the warmest desk to sit at in the building; this would require the building to utilize a dense deployment of temperature sensors. A third person in a chemical lab may want the building to monitor and notify him/her about the results of a chemical reaction that s/he started before leaving; this situation could likely be accomplished with a camera with special capabilities to detect these reactions. While it is possible to enable all these applications, these examples highlight three main challenges that prevent FMs and LLMs for achieving the same amount of autonomy in our physical environments as we have seen in the digital domain:

\noindent
\textbf{1. Each new application requires a new device or sensor, often with built-in special capabilities.} Each of the three examples we discussed would require the user to purchase or engineer a system to satisfy that task. Evolving new applications and functionality in this way is also not scalable. %

\noindent
\textbf{2. Events, tasks, and actions are often localized.} Events and tasks are often localized to small areas. As we will show in Section~\ref{sec:challenges_context}, an FM that is analyzing data from multiple sensors covering a large space often misses events occurring in small sections of a large scene. While LLMs and FMs have shown great performance in analyzing general trends, we will show that they have difficulty detecting and responding to localized events in larger sensor deployments (Section~\ref{sec:challenges_context}).

\noindent
\textbf{3. Achieving full coverage across an entire space requires space-dependent dense deployments.} The example of finding the ``best desk'' to sit requires a dense deployment of temperature sensors. While it is common for smart spaces to deploy smart devices and sensors, new applications may require dense deployments that need to be tailored to the layout of the environment, making it difficult to achieve full coverage for each new application.

We propose \newname, a platform that addresses the challenges in enabling FMs to interface with the physical world by enabling FMs, particularly visual-language models (VLMs), to ``zoom in'' to localized areas of interest to obtain a higher resolution, fine-grained, and better understanding of the physical environment to carry out the task at hand. \newname accomplishes this goal with two critical design choices, as shown in Figure~\ref{fig:teaser}. 

First, we propose a novel image segmentation method called Aspect Ratio Constrained K-Means (ARCK-Means) segmentation to visually ``zoom in'' and identify potential locations pertinent to the task. We show how existing state-of-art segmentation methods (e.g., Segment Anything Model (SAM)~\cite{kirillov2023segment}) often result in split objects that reduce object identification and localization accuracy. ARCK-Means segmentation ensures that full objects are processed and improves the analysis of localized areas of interest in FMs.

Second, we propose an adaptive, modular, and reconfigurable sensing and actuation drone platform that FMs can actuate to ``zoom in'' physically. Once the FM identifies potential locations of interest, it can select the relevant sensor or actuator to equip, before actuating the drone to analyze locations of interest up close. For example, to answer the question: ``where is the warmest place to sit?'', the FM would identify potential locations that may indicate warmth (e.g., sunlight), before sending a drone equipped with a temperature sensor to confirm. While there are several existing configurable drone platforms~\cite{schiano2022reconfigurable,derrouaoui2021nonlinear,perikleous2024novel,hert2023mrs}, these works focus on the physical design and control of the drone rather than the sensing, analysis, and actuation capabilities. Moreover, these \textit{existing works require manual reconfiguration, while \newname autonomously reconfigures sensing and actuation capabilities on-the-fly}. Our motivation for exploring the use of drones to empower foundation models interfacing with the physical world stems from our vision that \textbf{humans and robotic systems will coexist in the future.} Moreover, our design choices address the challenges mentioned previously:

\noindent
\textbf{1. Modularization enables easy integration of new applications.} Much like single-lens reflex (DSLR) cameras with interchangeable lenses, modularization 1) allows for the creation of an \textit{ecosystem} of sensors, actuators, and applications. Consumers can therefore purchase only the sensors and actuators for the applications they need and improve drone reusability, expendability, and sustainability. A static drone may not have all the required sensors and actuators, which would require purchasing a completely new drone. Layering and modularization also enables 2) evolving the drone and enabling new applications independently. New sensors and actuators purchased for new applications will still be compatible as long as the interface remains the same. Drones carrying a single sensor or actuator 3) can be designed much smaller, more \textbf{agile}, less noisy, and more suitable for closed environments, such as indoors.

\noindent
\textbf{2. Actuating a drone enables localized sensing and actuation.} A reconfigurable drone platform enables FMs to dynamically specify what sensing or actuation modality at which location, without relying entirely on static sensors that may not have been deployed.

\noindent
\textbf{3. Drones can achieve full spatial coverage while allowing for sparser deployments.} Because FMs can actuate a drone to any location, there no longer is a need for dense deployments for all types of sensors. Instead, a single drone can be used to service an entire space.

To demonstrate the utility of \newname for LLMs and FMs interfacing with physical environments, we prototype and show how \newname could be integrated into a \textit{personal assistant} system that leverages static sensors deployed throughout the environment (cameras) in conjunction with foundation models and penetrative AI~\cite{xu2024penetrative} to satisfy a wide range of useful tasks in a home, lab, or office setting. To the best of our knowledge, \textit{our work is the first to propose a drone platform with reconfigurable sensing to address challenges in coverage, localized sensing, and dense deployments for enabling general LLM and FM interactions with our physical environments}. Our contributions are as follows.

\noindent
\textbf{1.} We propose \newname, a platform that enables FMs, particularly VLMs and LLMs, to better understand and interface the physical world by identifying localized areas of interest from large scenes, equipping relevant sensors/actuators, and actuating the drone to ``zoom in'' up close.

\noindent
\textbf{2.} To realize \newname, 1) we propose a novel image segmentation method called ARCK-Means that aids FMs in identifying localized areas of interest. We demonstrate that ARCK-Means improves the understanding and detection of objects in large and cluttered scenes over existing segmentation techniques, namely SAM, by reducing the amount of disjoint and split objects produced by segmentation masks. 2) We introduce a drone-based modular and reconfigurable sensing and actuation platform that enables FMs to adapt to a wide range of scenarios. \newname actuates the drone, equipped with task-relevant sensors and actuators, to identify areas of interest to analyze up close.

\noindent
\textbf{3.} We demonstrate how \newname can augment LLMs and FMs and easily enable new applications throughout our environments, beyond the capabilities of common IoT smart devices, by prototyping a \textit{personal assistant} system that leverages both static sensors (cameras) and the mobility and flexibility of a reconfigurable drone. Our deployments and demo~\cite{zhao2024connecting} in realistic scenarios demonstrate that \newname can improve the success rate of a diverse array of tasks by up to $85\%$.

\section{Related Works} \label{sec:related}

\noindent
\textbf{1. Language and foundation models.} LLMs and FMs have seen a surge in usage and research due to their powerful capabilities in allowing computers to understand and interact using natural human modes of communication in an unprecedented manner. While most of these works focus on generating language, text, and digital media, there is a growing trend on adapting them for autonomous systems that interact with the physical world. Many such works target robotic platforms~\cite{vemprala2023chatgpt}, including drones and robot vacuums, that only have an egocentric view of the vicinity around them. Works that leverage FMs and LLMs for interacting with larger spaces, namely smart homes, generally focus on actuating common smart appliances to better respond to the needs of occupants~\cite{king2023get,nie2022ai,nie2022conversational}. These works, leverage LLMs to create a natural language interface between humans and their environments, often leveraging internet-connected devices and targetting applications that are already widespread (e.g., television or air conditioning control). Our work focuses on enabling new applications and interactions that LLMs and FMs can have with occupants and their environments scalably without being limited to the constraints imposed by the sensors and functionalities implemented and hard coded into the devices found throughout the environment.

\noindent
\textbf{2. Reconfigurable\add{ and adaptive} sensing \delete{and drone }platforms.} There are many reconfigurable and modular sensing platforms in existence. In addition to the platforms provided by open-source do-it-yourself (DIY) electronic vendors, such as Adafruit~\cite{adafruit-sensors} and Sparkfun~\cite{sparkfun-sensors}, there are also platforms that operate on even lower resource microcontrollers, without an operating system, that are less flexible in the number of interfaces and configurations they support~\cite{yi_2018}. ~\cite{zhao2022modular,zhao2023legosense} are reconfigurable sensing platforms based on the Raspberry Pi that have unified and generic hardware interfaces, allowing sensors to use the same set of connectors even if they are interfaced differently in software (e.g., UART, SPI, or I2C). 
Several platforms leverage the Berkeley TinyOS operating system~\cite{Levis_2005}, which allow developers to develop extremely long-lasting applications with great flexibility~\cite{culler2002mica, minaturized_2005,MASS_2005,miniaturised_2007, Mikhaylov_2013, Mikhaylov_2014}.

There are a few reconfigurable drone platforms, but most focus on enabling flexible physical design and control of drones~\cite{schiano2022reconfigurable,derrouaoui2021nonlinear,perikleous2024novel,hert2023mrs}. For example,~\cite{da2020drone} allows operators to reconfigure the number of rotors the drone with locking mechanisms that connect rotor modules. These works typically do not focus on sensing and actuation reconfiguration, unlike \newname. Additionally, they typically require a person to manually reconfigure the drone, while we introduce mechanisms that enables human-free configuration of the sensing and actuation. \cite{xia2021drone} recently proposed a reconfigurable drone platform, but is only a demo abstract and leaves out many details on how it autonomously swaps modules.

\add{Prior works have also explored adaptive sampling strategies to optimize sensor deployment and data collection, both in traditional wireless sensor networks~\cite{2011Pomili} and mobile sensing platforms like underwater vehicles~\cite{2012Pomili}. Unlike these approaches that focus on sampling optimization, \newname enables on-demand sensor reconfiguration guided by foundation models.}

\noindent
\add{\textbf{3. Embodied systems.} Traditional robotic systems often rely on hand-crafted algorithms, making adaptation to new or unforeseen situations challenging. Additionally, they struggle to generalize learned behaviors across different tasks, limiting their effectiveness in real-world applications~\cite{hu2023robofm}. By leveraging LLMs and FMs in robotics, including drones, the generalization problem in planning~\cite{huang2022planners, Lin_2023} and control~\cite{chen2024typefly, yang2025fm, singh2022promptrobot} in different tasks is partially resolved. However, these works still face severe limitations for solving diverse and general tasks due to insufficient real-world interactions. This is primarily because traditional robotic systems rely on limited and fixed sensing capabilities, often restricted to cameras; once a robotic system is built, their hardware capabilities are hardly ever upgraded. Many tasks require multiple sensing modalities or may need the use of different types of sensors to effectively understand, interpret, and respond to events in the physical world. To resolve the aforementioned limitations, we propose \newname, a system that bridges the physical world with foundation models through a modular sensing platform.
}

\section{Preliminaries and Challenges}
\label{sec:challenges_context}

In this section, we explore the limitations of current foundation models in understanding and responding to events and tasks that require interacting with physical environments. Such tasks a person may ask his/her home or environment could include ``where is the warmest place to sit'', ``where did I leave my phone'', ''bring me a snack'', and much more.

The growth of the Internet of Things (IoT) is exponential, and the number of internet connected sensors and devices is expected to double by 2030~\cite{sinha2024number}. It is expected that all future buildings and homes will be equipped with a wide array of different sensors and devices that will improve our efficiency, automation, personalized insights, and overall quality of life. We envision that foundation models will be capable of understanding data collected throughout our physical environments and act as a \textbf{natural interface between humans and their built environments}.

\subsection{Deployment}
\label{subsec:prelim_setup}

To test the current limitations of FMs for realizing this vision, we created the deployment shown in Figure~\ref{fig:localization_benchmark}e, using a network of 4 cameras. Since FMs for language and vision are most mature, we focus on these modalities; in the future, we believe that foundation models capable of interpreting streams of heterogeneous sensing modalities from sensors strewn throughout the environment will be possible.

We deploy a camera network on the ceiling of our deployment area, facing directly downwards. We generate floor maps and stitch together frames from individual camera views, according to~\cite{liu2022sofit}, which we input into the Large Language-and-Vision Assistant (LLaVA)~\cite{liu2023visual}, a state-of-art visual-language model (VLM), along with user specified tasks and commands. %

\subsection{Types of Tasks}
\label{subsec:prelim_task_types}

We identify and explore four classes of tasks in this work that are useful for creating smarter physical spaces, but are not commonly packaged into existing IoT smart devices:

\noindent
\textbf{T1: Object/Location Identification (ID).} This set of tasks requires the system to observe the environment and identify an object or location based on the user's command (e.g., ``where is my phone?''). In our preliminary analysis, we look at one task in this category: ``where is my phone?'' and place a phone at different locations in view of the cameras.

\noindent
\textbf{T2: Object/Location State (State).} This set of tasks involves learning about the state or condition of a specific object or location. For example, ``is my food burning'' would require the FM to identify food that is being cooked and if the food is producing too much smoke. In our preliminary analysis, we look at one task in this category: ``where is the warmest place to sit?'' and artificially turn up the heat in certain areas of the room by placing concealed space heaters nearby.

\noindent
\textbf{T3: Surveillance (Surv).} These tasks differ from the previous two, which are executed only once after the command is received. In surveillance tasks, the system needs to continuously analyze the environment until a potential event occurs (e.g., ``Let me know if any chemical in my experiment spills on the table''). In our preliminary analysis, we look at one task in this category: ``let me know if any chemical in my experiment spill on the table?'' and artificially simulate this event by knocking over a glass of colored water.

\noindent
\textbf{T4: Actuation (Act).} Unlike previous categories, which require sensing, actuation tasks require direct physical interactions with the environment (e.g., ``bring me my medicine'').

\subsection{Preliminary Analysis and Limitations}

\begin{figure}[t!]
    \centering
    \includegraphics[width=\linewidth]{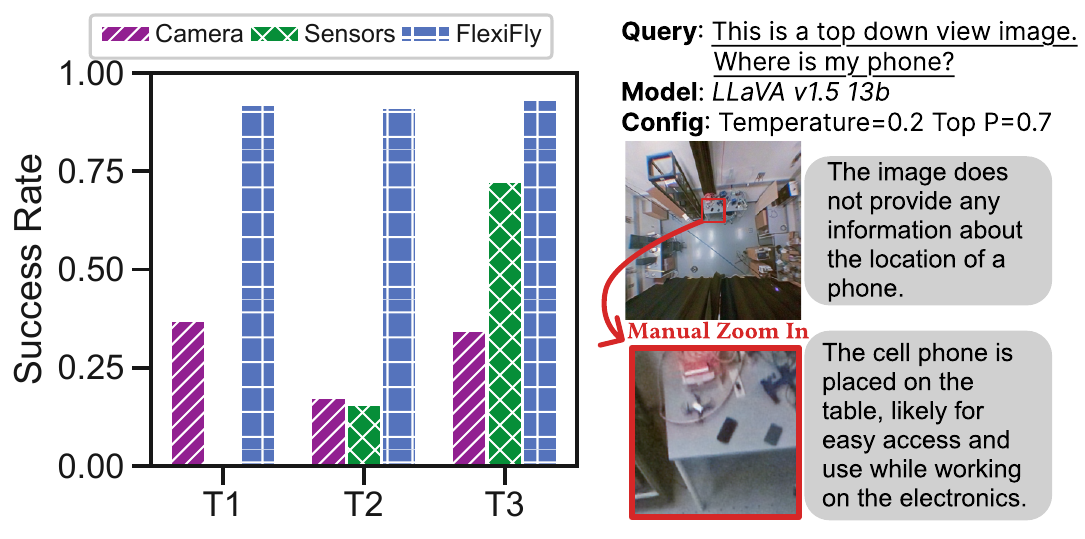}
    \caption{Preliminary study: (a) task completion rate of standard FMs leveraging VLM (camera) and dense sensor networks, compared to \newname; (b) Example showing that the VLM could not detect the phone in plain sight unless ``zoomed in''.}
    \label{fig:prelim_results}
\end{figure}

Using this basic camera network + FM setup (labeled VLM Baseline), we attempt to satisfy one command from each of the classes of actions we identified. We also compared against an FM analyzing data from a deployed sensor network  (labeled Sensor Baseline); here, we deployed 9 sensors (temperature for the state task and alcohol sensors for the surveillance task) in a grid pattern and interpolated between sensors to obtain a map spanning the entire space. We ran 70 trials for each scenario and show the success rate of identifying or completing each task in Figure~\ref{fig:prelim_results}a. We see that the success rate is low across the board for the VLM and sensor baseline, due to three limitations.

\noindent
\textbf{Limitation 1: Physical size of environments and volume of sensor information and is large, while events and objects of interest are localized to small areas.} A FM overseeing physical environments need to process many streams of data covering a large area, such as the space of our deployment. However, users are generally interested in only a small portion of the environment. We noticed that for the ID tasks (looking for phone) and surveillance (detecting spills), our FM could often not detect these objects and events due to the event occurring in a tiny portion of the environment and the limited resolution that can be captured. As shown in Figure~\ref{fig:prelim_results}b, LLaVA could not identify many objects unless we manually zoom in and reduce the camera's area of coverage.

\noindent
\textbf{Limitation 2: Coverage and resolution of sensing modalities required is limited and not practically scalable.} The VLM could not accurately identify the warmest place to sit (T2: state task) because it leverages a sensing modality where standard vision may not perform well (temperature). Even when we included temperature sensors (sensor baseline), the success rate is still low. When we simulated the warmest place to sit close to the sensor, the success rate was high, but the sensor network cannot accurately capture the dynamics of the space at areas away from the sensors, which is where success rate decreased. This highlights that the \textit{density of sensors the performance and understanding an FM has about the environment.} Additionally, while detecting the warmest place to sit may require a temperature sensor, \textit{different applications require different sensor deployments} (e.g., detecting falls could use audio or vibration). For instance if audio is used to detect falls and the bedroom does not have a smart speaker or microphone, then it would be difficult to enable this service in the bedroom.

\noindent
\textbf{Limitation 3: Actuation is restricted.} An actuation task such as bringing the user a snack or medicine cannot be accomplished with only an FM along and static devices throughout the environments.

\subsection{Design Philosophy}
\label{subsec:design_philosphy}

To address the limitations discussed previously, our design philosophy involves creating mechanisms that enable FMs to ``zoom in'' and sense targeted areas of interest with higher resolution. We tackle these limitations on two fronts. First, we propose a novel image segmentation technique called Aspect Ratio Contrained K-Means (ARCK-Means) to aid in identifying potential locations of interest to ``zoom in''. Second, we propose \newname, a drone platform, with modular and reconfigurable sensing and actuation, that allows FMs to pick and choose sensors and actuators depending on the task at hand. The FM first leverages ARCK-Means to identify areas of interest based on the current state of the environment and the input command (Section~\ref{sec:visual_llm}), before actuating the drone with the corresponding sensors to ``zoom in''(Section~\ref{sec:sensor_selection_landing}). 

For example, an FM looking to answer ``where is the warmest place to sit'' previously could only rely on analyzing and sifting through a large amount of sensor data from static sensors throughout the environment. With \newname, an FM identifies local areas of interest with ARCK-Means, before actuating a temperature sensor equipped drone to areas of interest to measure and compare temperature readings.

ARCK-Means alleviates the first limitation by aiding the FM in identifying local areas of interest in large cluttered scenes. Our reconfigurable sensing and actuation drone platform alleviates limitation 2 and 3. Instead of requiring a dense deployment of static sensors (limitation 2), FMs can reconfigure and actuate the drone to the precise location it needs to sense, which reduces deployment overhead. Moreover, a mobile drone platform enables actuation of items in the physical environment that are more restrictive for static sensors and devices (limitation 3).

\section{Identifying Local Areas of Interest via Segmentation.}
\label{sec:visual_llm}

\begin{figure}[t!]
    \centering
    \includegraphics[width=\linewidth]{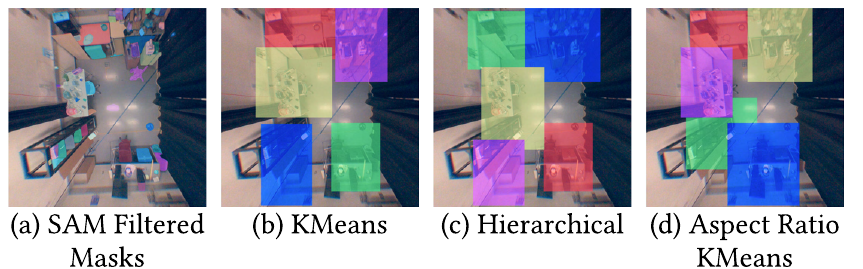}
    \caption{Segmentation and clustering to break down scenes into smaller more manageable pieces for LLaVA and DINO. (a) Object masks after applying Segment Anything Model (SAM); (b) Extracted frames after clustering object masks based on K-Means; (c) Hierarchical clustering, and (d) ARCK-Means. For ARCK-Means, we constrain the aspect ratio of extracted frames to be between 0.67 and 1.5.}
    \label{fig:clustering}
\end{figure}

\begin{figure*}%
  \begin{center}
    \includegraphics[width=\textwidth]{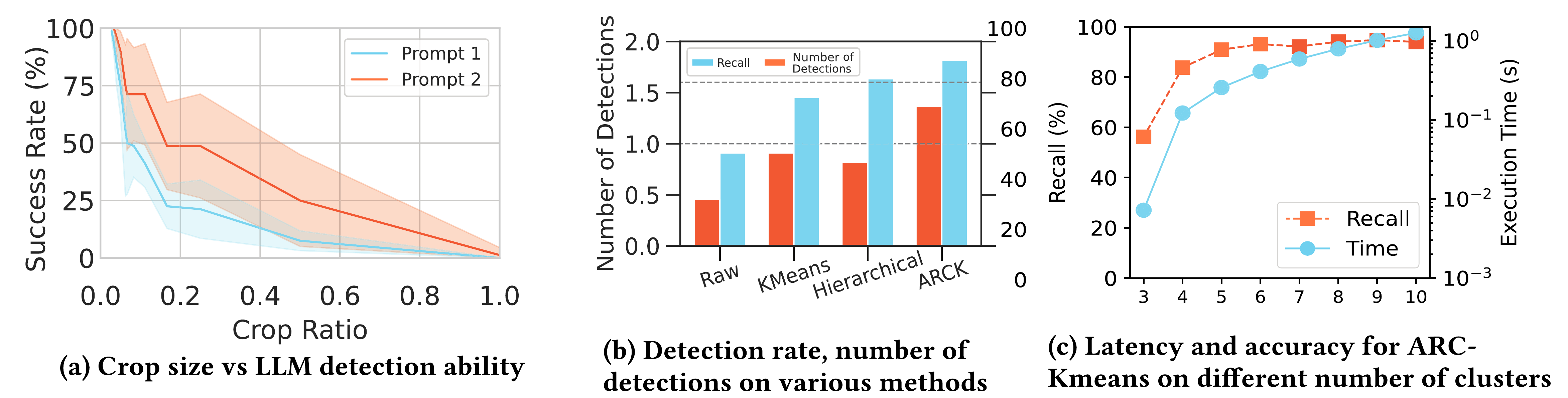}
  \end{center}
  \caption{Different clustering methods and thresholding evaluated for segmentation. Prompt 1: \textit{Describe the image in detail.} Prompt 2: \textit{Is there a \{object name\} in the image?}}
  \label{fig:clustering_benchmarking}
\end{figure*}

As mentioned in Section~\ref{sec:challenges_context}, when a user command arrives, we input the command along with frames stitched together from the camera network in LLaVA. We found that LLaVA was not able to reliably detect objects of interest in a scene when we passed in an image or a large space (e.g., a single chair in a large room); the larger the scene, the less detailed and more high-level the responses became.

One way to alleviate this is to segment and analyze smaller portions of the full image. Instead of directly passing the entire scene into LLaVA, we use the state-of-art image segmentation model, Segment Anything Model (SAM), to extract key objects and smaller areas of the scene~\cite{kirillov2023segment}. Figure~\ref{fig:clustering}a shows an example of the object masks that are output by SAM. These object masks are then input into LLaVA, along with the user command, and outputs (in text) the names of potential objects in the scene that might be of interest. These text outputs from LLaVA and the segmented images are input into Grounding DINO~\cite{liu2023grounding}, a state-of-art language model for zero-shot object detection and localization, which then outputs localized bounding boxes of potential objects of interest. In implementing this segmentation pipeline for identifying locations of interest, we noticed a number of limitations that impacted performance, as discussed next.

\subsection{Limitations of Current Segmentation Methods}

\noindent
\textbf{1. SAM often generates segmentation masks that split objects between two different masks.} As such, LLaVA and Grounding DINO often identify the same object multiple times at slightly different locations when directly using the masks from SAM, adding to the processing time.

\noindent{2. The segmented masks are not rectangular, which reduces object detection performance.} We observed that directly inputting the segmented masks into LLaVA and Grounding DINO saw a large performance degradation. This performance degradation persisted even if we zero pad the segmented images into a rectangular shape. We suspect this is because LLaVA and Grounding DINO are typically trained with rectangular images, with standard aspect ratios (e.g., 640 x 480), in natural settings; whereas, the inputs we were trying are non-rectangular with most of the background removed (zero padded).

\subsection{Aspect Ratio Constrained K-Means (ARCK-Means) Segmentation}

To address the limitations of current segmentation methods in aiding FMs identify potential locations of interest, we propose Aspect Ratio Constrained K-Means (ARCK-Means) Segmentation. ARCK-Means operates on the segmented objects generated by SAM.

To alleviate the challenge of splitting objects between different masks (limitation 1), ARCK-Means first clusters the centroids of the masks generated into $k$ clusters. We leverage hierarchical clustering~\cite{murtagh2012algorithms}, but also benchmark against K-means~\cite{ahmed2020k}. 

To ensure that whole rectangular images are analyzed by LLaVA and Grounding DINO (limitation 2), we find the minimum-area rectangle that encompasses all masks of each cluster. Moreover, we constrain the aspect ratio of clustered masks between to ensure aspect ratios of typical images used during training, resulting in the full ARCK-Means method. To accomplish this, we check if adding a new segment into the existing cluster causes the aspect ratio to fall below or above our constraints; if these constraints are broken, then we do not make the assignment. The resulting image segments are then analyzed by LLaVA and DINO.

\subsection{Analysis and Benchmarking}

Figure~\ref{fig:clustering}b-d shows an example of segmentation and clustering. We see that for ARCK-Means the clustered masks generated tend to be more square. Additionally, the cluttered table in the bottom right hand corner is fully encompassed by ARCK-Means, but both K-Means and hierarchical clustering do not capture all of the items on the table in a single segment. These improvements of ARCK-Means over the other clustering methods yields higher performance in detecting objects of interest, as shown in Figure~\ref{fig:clustering_benchmarking}. In this case and for the rest of the paper, a successful ``\underline{recall}'' means that the Grounding DINO model was able to detect and localize the object interest in any one of the segmented clusters it was given, regardless of any additional detections.

First, we tested two prompts with the LLaVA model and vary the size of the object with respect to the frame of one camera (Figure~\ref{fig:clustering_benchmarking}a). The first prompt is more \textit{general}, asking LLaVA to describe the objects present. The second prompt is more specific, asking if the \textit{specific} object of interest is present in the scene. We see that the more specific a prompt is, the higher the recall, which is the reason why we used more specific prompting. Second, we see that as the size of the object gets smaller with respect to the image frame, the recall gets smaller, since the signal-to-noise ratio of the object gets smaller. %

In Figure~\ref{fig:clustering_benchmarking}b, we compare the recall of each clustering method after clustering each scene into 5 segments, and see that ARCK-Means has the highest recall due to improvements in the masks it clusters (addressing limitation 1) and the shape of the resulting segments (addressing limitation 2); as such, we adopt ARCK-Means into the final system. Figure~\ref{fig:clustering_benchmarking}c shows the run time and recall of ARCK-Means as a function of the number of clusters or segments we split the scene into. We see that the recall levels out at around $90\%$ after $5$ clusters, while taking around $100$ms to run the full pipeline. Adding in more clusters does not yield significant improvements in detection, but significantly increases run time; as such, we segment the scene into five segments in the final system. To generate these plots, we used 47 images of indoor home, office, and lab environments. We took 35 images from the ADE20K scene parsing dataset~\cite{zhou2017scene} as well as 12 images from our own deployment. We implement and run the full visual-language model pipeline (SAM, clustering, LLaVA, and DINO) on an Nvidia RTX 3090 GPU server.

\section{Modular and Reconfigurable Sensing and Actuation Drone} \label{sec:sensor_selection_landing}

In this section, we introduce the reconfigurable and modular sensing and actuation platform for drones, which FMs can actuate to identified locations of interest (Section~\ref{sec:visual_llm}).

\begin{figure*}[t!]
  \begin{center}
    \includegraphics[width=\linewidth]{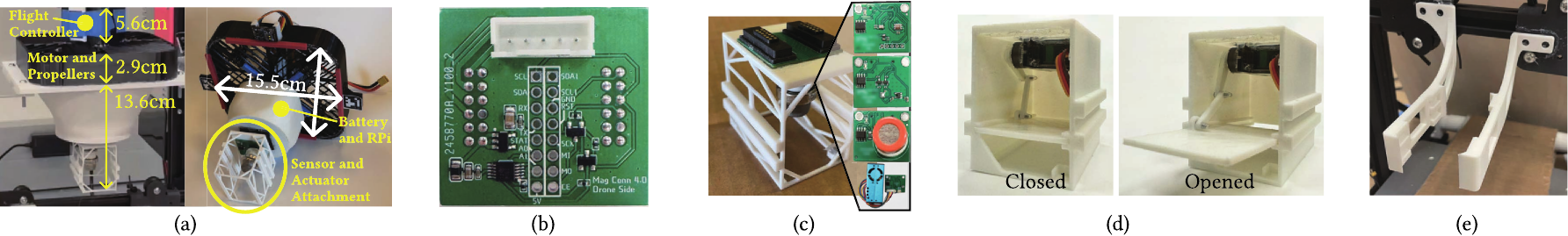}
  \end{center}
  \caption{a) Physical drone system, b) Carrier board, c) Several sensor modules, d) Actuation module, e) Gripper for swapping sensor and actuation modules}
  \label{fig:drone_system}
\end{figure*}
\subsection{Challenges and Design}
\label{subsec:challenges_and_design_drone}

\begin{figure*}%
  \begin{center}
    \includegraphics[width=\textwidth]{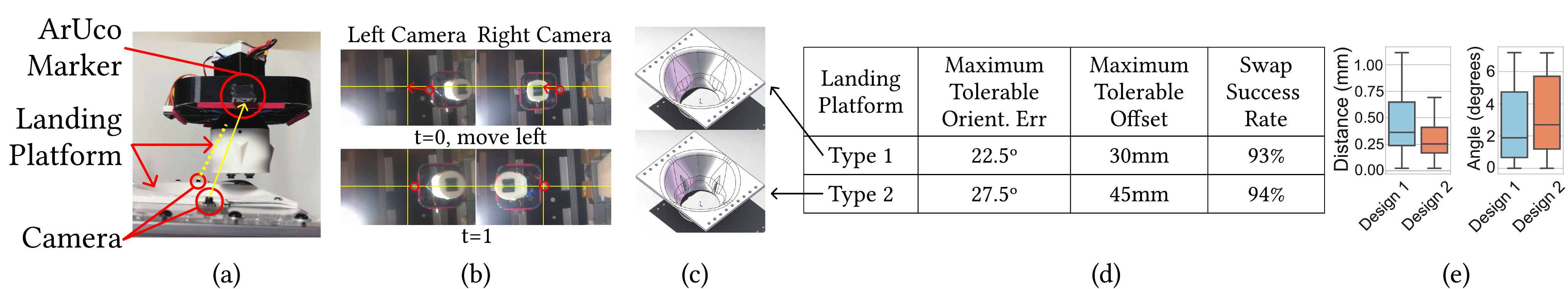}
  \end{center}
  \caption{Different landing platforms designed and benchmarked. The average landing time took $7.8$ seconds.}
  \label{fig:takeoff_landing}
\end{figure*}

The primary challenges in realizing an automated and reconfigurable sensing platform for drones is two fold.

\noindent
\textbf{1. Drone and module connector.} To allow reconfiguration, the sensor and actuation modules need to be detachable from the drone's main body. While there are drone platforms that can be reconfigured with different frames, wings, or motors~\cite{derrouaoui2022comprehensive,schiano2022reconfigurable,da2020drone}, the vast majority leverage mechanical connectors with locking mechanisms that often requires additional force and complex mechanisms to fasten and remove (e.g., grippers with joints much like human hands). Instead, \textit{we leverage a fully magnetic connector}. The advantage of this design choice is two-fold.

First, it simplifies the design of the mechanisms for removing and fastening new modules on the mechanical layer. Second, less force is required to fasten and remove the module. Compared to a mechanical design, such as~\cite{da2020drone}, that requires the application of force with several mechanisms at multiple locations, potentially damaging the module, connector, or drone, a magnetic connector enables gripper to bring a new module within vicinity of the drone before the magnet automatically aligns the pins and fastens the module in place. Removing the module follows a similar procedure. \emph{We tested the success rate in swapping a module on a drone} with both magnetic~\cite{amazon2024magnetic} and standard PCB mezzanine connectors~\cite{amp-connector-plug,amp-connector-receptacle} using our gripper (Figure~\ref{fig:drone_system}e) that actuates vertically to bring modules up and down. We ran 10 trials and found that our system could swap in new modules successfully every time when the connectors are magnetic. However, our system failed at swapping modules with the standard mezzanine connectors because our mechanical grippers could not apply enough force to remove nor attach modules.

\noindent
\textbf{2. Alignment.} Swapping in new modules onto the drone can only occur successfully if the drone is oriented correctly with the gripper after landing. This challenge is not directly related to the well studied problem of drone landing~\cite{wang2022micnest}, which aims to guide the drone to the landing platform before descending. As the drone descends closer to the ground, nonlinearities in drone stability (ground effect)~\cite{matus2021ground}, for which there is no promising method to compensate, can cause the drone to become misaligned. Instead, we focus on the design of the landing platform that automatically calibrates the orientation of the drone as it touches down.

Rather than creating a flat landing pad that is similar to the landing pads for helicopters, we take inspiration from the Ring Always Home Drone, which uses a funnel~\cite{ringalwayshome}. As long as the drone lands within the opening of the funnel, the drone will automatically slide towards the bottom of the funnel with an opening that latches onto the onboard module (Figure~\ref{fig:takeoff_landing}c-top). To further improve alignment, \textit{we propose a new design that includes additional grooves} to further improve module and drone alignment (Figure~\ref{fig:takeoff_landing}c-bottom).

We benchmark both the vanilla funnel and the grooved funnel landing platform design, as shown in Figure~\ref{fig:takeoff_landing}d. For each design, we had the drone land and swap modules 75 times and found that the swap rates are fairly similar. However, the grooved design corrects for greater drone misalignments from the platform, as reflected in the higher maximum tolerable orientation and offset errors. As such, we adopt our proposed grooved funnel design into \newname.

\subsection{Platform Components and Implementation Details}
\label{subsec:platform_components}

This section discusses the implementation of the rest of the reconfigurable platform.

\noindent
\textbf{Drone platform.} We build \newname off the open-source Crazyflie drone~\cite{crazyflie}, as shown in Figure~\ref{fig:drone_system}. We replaced its brushed motors with more powerful brushless motors (3800 Kv powered by a 4-cell 850mAh battery) \update{and 3D printed a guard to increase payload and improve safety. }{. To improve payload capacity and ensure operational safety in areas with human presence, we designed and fabricated a enclosed drone frame using lightweight foaming polylactic acid (LW-PLA), as shown in Figure \ref{fig:drone_system}a.}

\noindent
\textbf{Carrier board.} The carrier board, attached to the base of the drone, provides the physical data, power, and communication connections between the drone platform and each sensor/actuation module. We implement the carrier board on a lightweight \$15 Raspberry Pi Zero 2W~\cite{rpi02w}, which has magnetic connections for one sensor or actuation module.

\begin{table}
    \centering
    \resizebox{0.9\columnwidth}{!}{%
    \begin{tabular}{|c|c|c|c|c|c|c|} \hline 
        & Total& Module& Fly& Drone& Module& Cost\\
               & Mass& Mass& Time&             Power&              Power& \\ \hline 
        Drone Only & 344.7g& -& 3m47s& 195.4W& -& \$344\\ \hline 
        PM2.5& 411.9g& 67.2g& 3m16s& 226.0W& 0.29W& \$40\\ \hline 
        Temp\&Moisture& 372.3g& 27.6g& 3m43s& 198.8W& 3.3$\mu$W& \$12\\\hline
 Light Sensor& 372.8g& 28.1g& 3m43s& 199.3W& 10$m$W&\$10\\\hline
 CO$_2$& 372.8g& 28.1g& 3m42s& 199.8W& 86$m$W&\$16\\ \hline 
        Alcohol& 376.2g& 31.5g& 3m39s& 201.7W& 0.75W& \$8\\ \hline 
        Actuator& 394.8g& 50.1g& 3m08s& 235.1W& 1.22W& \$7\\ \hline
    \end{tabular}
    }
    \caption{\newname's supported sensors/actuators and their mass, power consumption, and cost.}
    \label{tab:sensor_actuator_list_mass_power_cost}
\end{table}

\noindent
\textbf{Sensor and Actuation Modules.} \newname comes with a collection of sensor and actuation modules (full list in Table~\ref{tab:sensor_actuator_list_mass_power_cost}). These modules are attached or detached from the drone by the ground station and comes with a 3D printed structure that increases the surface for the ground station to pick up modules. The \textit{actuation module}, shown in Figure~\ref{fig:drone_system}d, is a container structure with a motor that opens and closes a hatch. Small items (e.g., medication, candy, pet food, etc.) can be loaded into this module for the drone to deliver.

\noindent
\textbf{Ground station and automated takeoff and landing.} The ground station consists of a gripper mechanism and a conveyor belt to position modules below the drone. We created the platform leveraging the chassis of the open-source Ender-3 3D printer~\cite{ender3github}. The gripper in Figure~\ref{fig:drone_system}e removes and attaches modules onto the drone.

To land, the ground station guides the drone using two cameras facing up on top (Figure~\ref{fig:takeoff_landing}b). We also print and attach two \textit{ArUco} markers to the bottom side of the drone to easily detect the position of the drone and its orientation. ArUco markers, commonly used for camera pose estimation, are similar to QR codes, but carry less encoded information, which makes them more computationally efficient to detect~\cite{sani2017automatic}. In future work, we plan to leverage more complex computer vision models to automatically determine the position and orientation of the drone without needing to add additional markings.

\section{Implementation of a Foundation Model and Drone-based Assistant}

We show how FMs can leverage our novel methods for identifying local areas of interest (Section~\ref{sec:visual_llm}) and reconfigurable drone platforms (Section~\ref{sec:sensor_selection_landing}) to better understand and interact with physical environments, with the implementation and evaluation of a personal assistant system.

Figure~\ref{fig:personal_assistant_architecture} shows the workflow of our prototype. Static cameras in the environment provide a high-level and coarse-grained view of the environment, just like in our preliminary deployment (Section~\ref{subsec:prelim_setup}). When a user gives a voice command, the system leverages LLaVA and ARCK-Means segmentation (Section~\ref{sec:visual_llm}) to identify local areas of interest and the relevant sensor or actuator. Then, the system actuates the \newname drone with the sensor/actuator to each location to complete the task. We carry out and evaluate this system on the four classes of tasks we identify in Section~\ref{subsec:prelim_task_types}.

\noindent
\textbf{Implementation Details.} \add{A local server runs the Ollama framework, hosts the LLMs (Llama-3.1-8B), VLM (LLaVA 1.6-8b), and Grounding DINO open-set object detection model. Upon user request, a snapshot is taken from the ceiling camera network and stitched into a single image. The server identifies required sensing modules and key areas of interest, manages pipeline execution, and sends commands to the drone. A validation LLM (Llama-3.1-8B) is used that ensures properly formatted outputs throughout the process.} \update{In the case of a sensing task}{For sensing tasks} (ID or State tasks from Section~\ref{subsec:prelim_task_types}), the drone flies to each location while reading sensor values from the attached module. \update{After the drone lands back at the ground station, the time series data is input into GPT-4, a state-of-art language model~\cite{gpt42023openai}, to analyze and make the final determination. To enhance reasoning and decision-making, we employ Chain of Thought (CoT) prompting when querying the LLMs. If the sensing modality is to use a camera to identify an object (e.g., a objection/location ID task); the image from the drone is input back into the Grounding DINO model to make the final determination.}{If the sensing modality involves camera-based object identification (e.g., an object or location ID task), the captured drone image is processed on our local server. However, for time-series data analysis, the system uses GPT-4~\cite{gpt42023openai}, as we found that locally hosted LLMs often produce unreliable results when generating code to analyze the sensor data.} In an actuation task, the system attaches the actuation module with the relevant payload (e.g., snack or medicine) before flying to the destination for dropoff.

\noindent
\textbf{Prompting users for more context.} Some phrases require more context to properly identify locations, sensors, or actuators. For example ``tell me the `best' location to sit''. The word ``best'' could have many meanings (e.g., warmest, coolest, quietest, etc.). If a user gives a command with a non-specific adjective, such as ``best'', the system will prompt the user to clarify and be more specific. In the case of an actuation task, the system aims to deliver its payload to one location. If the visual-language model detects multiple potential locations, it will ask the user to clarify the location, either through voice or a web application that we implemented.

\subsection{Drone Navigation}
\label{subsec:drone_navigation}

\begin{figure*}%
  \begin{center}
    \includegraphics[width=\textwidth]{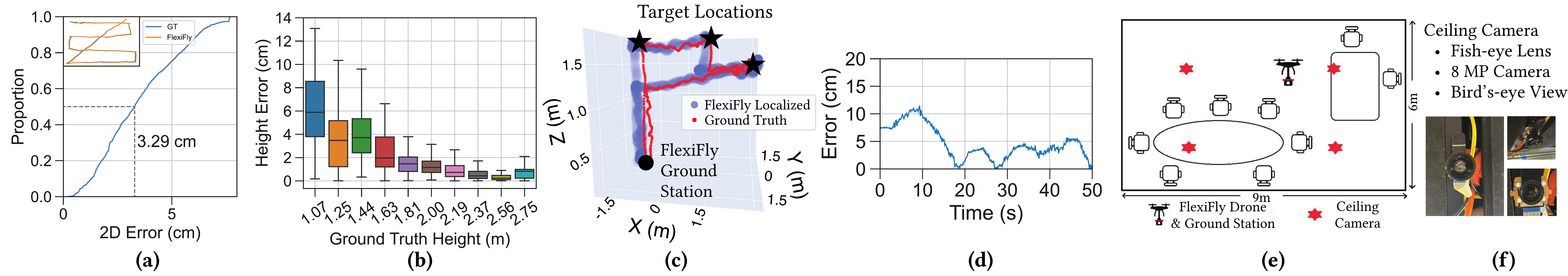}
  \end{center}
  \caption{a) 2-D drone localizaton CDF. b) Height localization error when drone is at different heights. c) Example ground truth and localized path of drone. d) Localization error of drone vs. length of mission. The localization error remains relatively constant, even as the time of the mission gets longer, demonstrating that the system is not susceptible to localization error drifts. e) deployment floormap. f) camera module used in ceiling camera network.}
  \label{fig:localization_benchmark}
\end{figure*}

During flight, we use images from the camera network to guide the drone by attaching an ArUco marker to the top of the drone, just like the patterns used for landing the drone in Section~\ref{sec:sensor_selection_landing}. To move the drone to specific locations, we use the straight line path from the drone's current location to the closest point of interest. Throughout our deployments in Section~\ref{sec:deployment}, we observed an median 2-D localization error of 3.29cm, using our camera network. Moreover, as shown in Figure~\ref{fig:localization_benchmark}, the localization error does not increase over time, as is common in inertial measurement unit and dead reckoning approaches. While there are more practical approaches for drone navigation, the focus of this work is to demonstrate the utility of \newname to LLMs that interact with the physical world. Hence, we leave these aspects for future work.

\section{Deployment and Evaluation} 
\label{sec:deployment}

We deployed our system into an office/lab setting as shown in Figure~\ref{fig:localization_benchmark}e, just as in our preliminary study. The goal of this deployment is to demonstrate improvements in task completion rate \newname provides to FMs for more general and less structured applications.

\subsection{Benchmarking}
\label{subsec:deployment_lab}

\begin{figure*}[t!]
  \centering
  \includegraphics[width=0.22\textwidth]{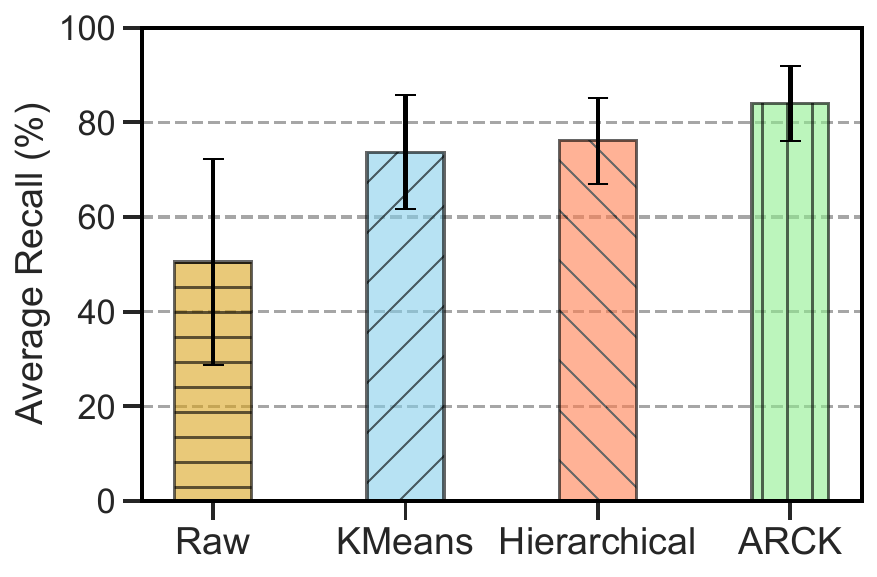}
  \hspace*{\fill} %
  \includegraphics[width=0.25\textwidth]{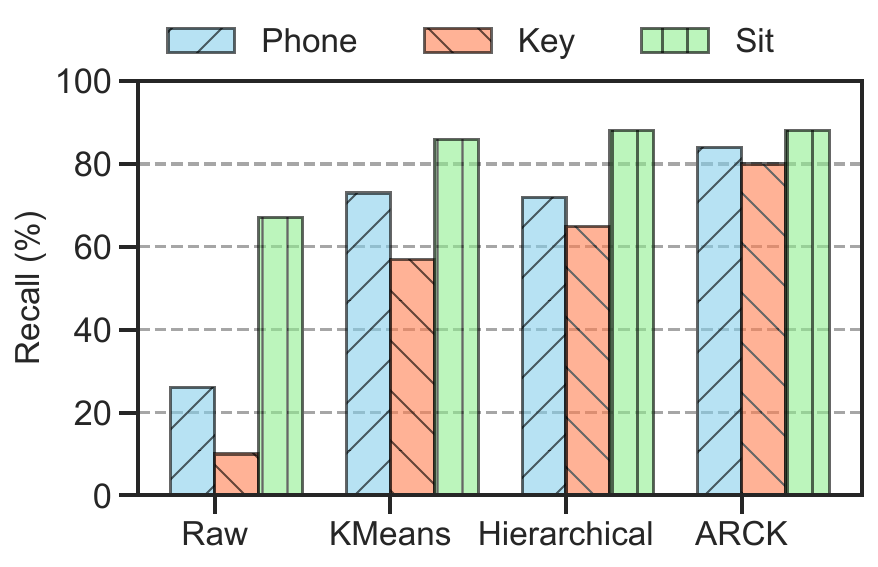}
  \hspace*{\fill} %
  \includegraphics[width=0.25\textwidth]{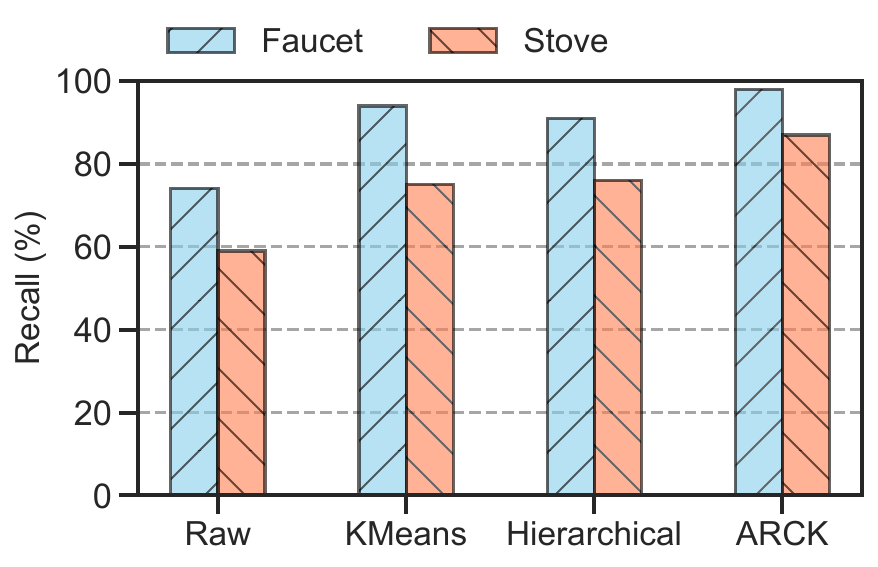}
  \hspace*{\fill} %
  \includegraphics[width=0.25\textwidth]{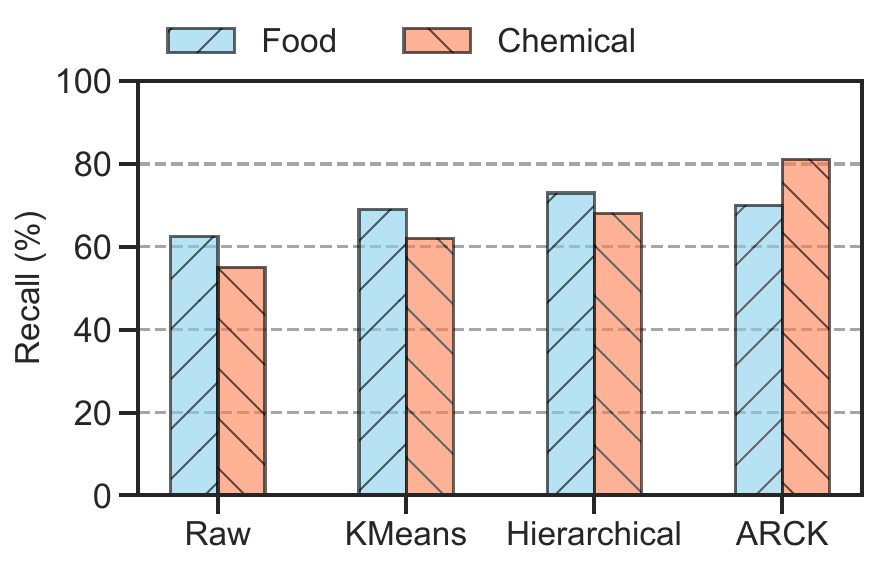}
  \caption{Summary of object and event detection by the LLaVA + DINO visual-language pipeline averaged (a) and broken down by category of sensing tasks (b-d). %
  }
    \label{fig:visual_llm_benchmark_summary}
\end{figure*}

\begin{figure*}[t]
\centering %
\begin{tabular}{ccc}
    \includegraphics[width=0.32\textwidth]{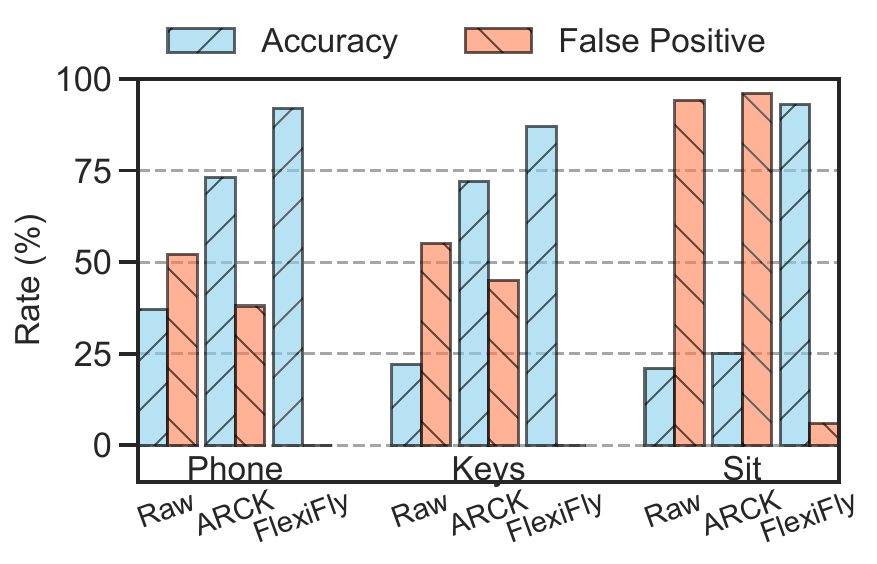} &
    \includegraphics[width=0.32\textwidth]{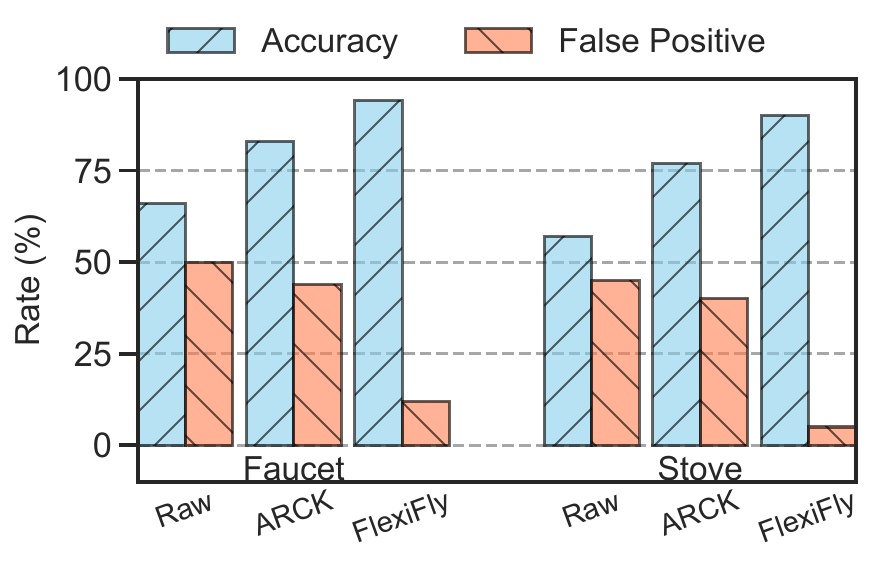} & 
    \includegraphics[width=0.32\textwidth]{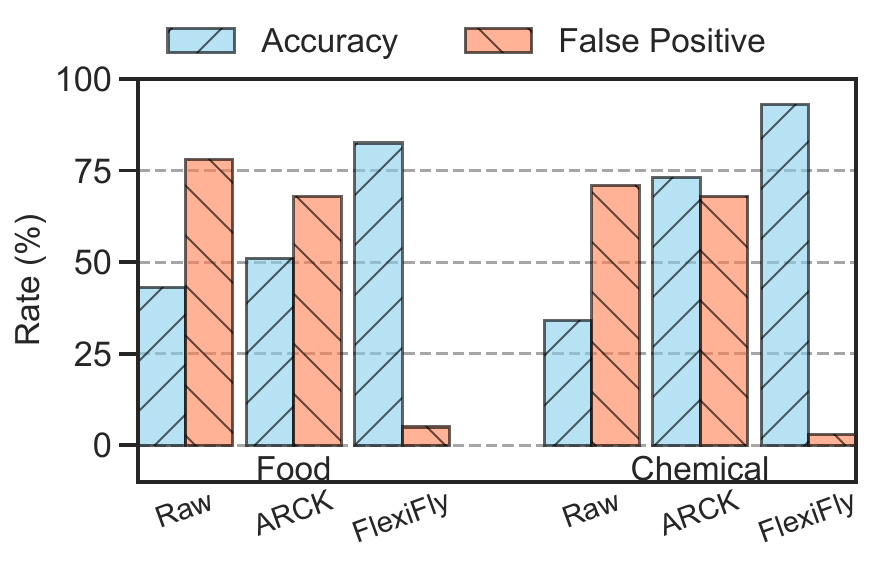}   
\end{tabular}
\caption{Breakdown of accuracy and false positive detections by sensing task category with and without \newname. We see that leveraging \newname in conjunction with static cameras greatly reduces false detections and improves accuracy because the drone can get a closeup view or sense an important part of the environment that a camera alone cannot (e.g., humidity).} 
\label{fig:accuracy_fp_benchmark_realworld} 
\end{figure*}

\begin{table*}
    \centering
    \resizebox{\linewidth}{!}{%
    \begin{tabular}{|l|c|c|c|c||c|c|c|c||c|l|c|c|c|c|} \hline 
         & \multicolumn{4}{|c||}{Camera Baseline} & \multicolumn{4}{|c||}{Camera + ARCK-Means only (Section~\ref{sec:visual_llm})}&   \multicolumn{5}{|c|}{Camera + \textbf{\newname} (Sections~\ref{sec:visual_llm} and~\ref{sec:sensor_selection_landing})}\\ \hline 
         Scenario& Precision&  Recall&  F-1&  Accuracy& Precision&  Recall&  F-1&  Accuracy&   Sensor Used&Precision&  Recall&  F-1& Accuracy\\ \hline 
          \multicolumn{14}{|l|}{\textbf{Object / Location Identification}}\\     
         \hline 
         Find Phone& 33.33\% & 26.00\% & 29.21\% & 37.00\% & 68.85\%&  84.00\%&  75.68\%&  73.00\%&   Drone Cam&100.00\%&  84.00\%&  91.30\%& 92.00\%\\ \hline 
         Find Key& 26.67\% & 10.00\% & 14.55\% & 21.67\% & 78.05\%&  80.00\%&  79.01\%&  71.67\%&   Drone Cam&100.00\%&  80.00\%&  88.89\%& 86.67\%\\ \hline 
         Sit - Temperature& 15.91\% & 56.00\% & 24.78\% & 17.48\% & 23.47\%&  92.00\%&  37.40\%&  25.24\%&   Temperature&76.67\%&  92.00\%&  83.64\%& 91.26\%\\ \hline 
         Sit - Humidity& 20.45\% & 66.67\% & 31.30\% & 21.00\% & 25.81\%&  88.89\%&  40.00\%&  28.00\%&   Humidity&82.76\%&  88.89\%&  85.71\%& 92.00\%\\ \hline 
         Sit - Light& 20.88\% & 79.17\% & 33.04\% & 25.96\% & 20.62\%&  83.33\%&  33.06\%&  22.12\%&   Light Sensor&95.24\%&  83.33\%&  88.89\%& 95.19\%\\ \hline 
         Average (ID)& 23.45\% & 47.57\% & 26.58\% & 24.62\% & 43.36\%&  85.64\%&  53.03\%&  44.01\%&   &90.93\%&  85.64\%&  87.69\%& 91.42\%\\ \hline 
          \multicolumn{14}{|l|}{\textbf{State of Object / Location}}\\ \hline 
 Faucet Open& 72.83\% & 74.44\% & 73.63\% & 65.71\% & 80.00\%& 97.78\%& 88.00\%& 82.86\%&  Humidity&93.62\%& 97.78\%& 95.65\%&94.29\%\\ \hline 
 Stove Open& 69.49\% & 58.57\% & 63.57\% & 57.27\% & 79.22\%& 87.14\%& 82.99\%& 77.27\%&  Temperature&96.83\%& 87.14\%& 91.73\%&90.00\%\\ \hline 
 Average (State)& 71.16\% & 66.51\% & 68.60\% & 61.49\% & 79.61\%& 92.46\%& 85.50\%& 80.06\%&  &95.22\%& 92.46\%& 93.69\%&92.14\%\\ \hline 
  \multicolumn{14}{|l|}{\textbf{Surveillance}}\\ \hline 
 Food Burning& 44.64\% & 62.50\% & 52.08\% & 42.50\% & 50.91\%& 70.00\%& 58.95\%& 51.25\%&  PM&93.33\%& 70.00\%& 80.00\%&82.50\%\\ \hline 
 Chemical Spill& 18.03\% & 55.00\% & 27.16\% & 34.44\% & 25.40\%& 80.00\%& 38.55\%& 43.33\%&  Gas (Alcohol)&88.89\%& 80.00\%& 84.21\%&93.33\%\\ \hline 
 Average (Sur.)& 31.34\% & 58.75\% & 39.62\% & 38.47\% & 38.15\%& 75.00\%& 48.75\%& 47.29\%&  &91.11\%& 75.00\%& 82.11\%&87.91\%\\ \hline 
  \multicolumn{14}{|c|}{}\\ \hline 
 Average (all)& 31.18\% & 51.74\% & 34.46\% & 32.17\% & 46.54\%& 83.17\%& 55.70\%& 48.99\%&  &91.93\%& 84.79\%& 87.78\%&90.80\%\\ \hline
        \end{tabular}
    }
    \caption{Summary of end-to-end performance between \newname and camera-only for all sensing tasks.}
    \label{tab:summary_performance_tables}
\end{table*}

We ran 2-3 tasks in each of the categories of tasks, as we discuss next. For each task, we issued 70 different trials. The scenarios are described in more detail next.

\noindent
\textbf{1) Phone.} In this ID task, the user asks ``where is my phone''. The system will actuate the drone with a camera module to potential locations to detect the phone. 

\noindent
\textbf{2) Key.} This ID task is similar to the \textit{phone} task, but instead users are looking for keys.

\noindent
\textbf{3) Sit - X.} In this series of ID tasks, users ask ``where is the best place to sit'', based on some sensing modality (e.g., `sit - temp' is where the user asks for the coolest or warmest place). We artificially increase or lower temperatures at different seats by placing space heaters or fans nearby.

\noindent
\textbf{4) Faucet.} In this state task, users ask ``is my faucet still on''; the system will then actuate the drone with a moisture sensor to detect the presence of large amounts of water leaking.

\noindent
\textbf{5) Stove.} In this state task, users ask ``is my stove still on''; the system will then actuate the drone with a temperature sensor to detect the state of the stove.

\noindent
\textbf{6) Food.} In this surveillance task, users ask ``let me know when my food is burning''. We simulate burning food by boiling water. The system will then actuate the drone attached with a particulate matter sensor.

\noindent
\textbf{7) Chemical.} In this surveillance task, users ask ``let me know if any chemicals spill''. We will then knock down and spill a glass of alcohol. The system will then actuate the drone with an alcohol sensor to confirm.

\noindent
\textbf{8) Medicine.} In this actuation task, the user will ask ``please bring me my medicine'' and wave his/her arms at a camera above. The system will then attach an actuation module loaded with vitamins on the drone, which will then deliver it to the person.

\noindent
\textbf{9) Poison.} In this actuation task, the user will direct the drone to ``deliver rat poison'' to a specific location in the environment. The system will load an actuation module with rat poison pellets (simulated with small snacks) and the drone will dispense them at the specified location.

\subsection{Results and Analysis}

We compare \newname against two baselines: camera + LLaVA setup from our preliminary study (Camera Baseline) and augmenting the camera baseline with ARCK-Means only (Camera + ARCK-Means). For non-actuation tasks, Figure~\ref{fig:visual_llm_benchmark_summary} shows a breakdown of the visual-language model performance in identifying the locations to send the drone versus the clustering method (Section~\ref{sec:visual_llm}). We see that the ARCK-Means clustering method yields the highest recall, across all scenarios, due to improvements in maintaining common aspect ratios and whole objects over other methods.

Figure~\ref{fig:accuracy_fp_benchmark_realworld} highlights improvements that \newname brings compared to a purely static camera-based system across all sensing tasks. A successful or ``accurate'' trial in this context means that the system was able to correctly identify the correct object or location (object/location ID task), correctly identify the state of the object or location (object/location state task), or correctly identify when a targeted event occurs (surveillance task); any additional points of interest identified are counted as incorrect identifications (false positives). Because we are often looking for small item(s) and locations in a large scene, the visual-language model pipeline often identifies multiple points of interest (e.g., DINO draws multiple bounding boxes and locations). Without a platform such as a drone that can ``zoom in'' and confirm, the sensing capabilities of this system is limited, and the false positive rate becomes extremely high, and the precision becomes low with additional locations identified. However, adding in the \newname-equipped drone allows the system to actuate the drone to each location to obtain a closeup view of the location and remove extraneous locations or sense an aspect of the environment that a camera cannot (e.g., humidity). This both reduces false positives and improves overall accuracy. Table~\ref{tab:summary_performance_tables} breaks down the recall, precision, and f-1 score across all individual tasks to further illustrate improvements in true detection rate (recall). In total, integrating \newname with FMs improved the task success rate by $85\%$.

\begin{table}
    \centering
    \resizebox{\columnwidth}{!}{%
    \begin{tabular}{|l|c|c|c|} \hline 
         Scenario&  \# of user prompts&  \# of executions&  Execution\\
          &per execution&per battery&Time\\ \hline 
 \multicolumn{4}{|l|}{\textbf{Object / Location Identification}}\\\hline     
         Find Phone&  1.0&  7&  44.4s\\ \hline 
         Find Key&  1.0&  7&  46.0s\\ \hline 
         Sit - Temperature&  2.6&  3&  84.3s\\ \hline 
         Sit - Humidity&  1.9&  3&  87.7s\\ \hline 
         Sit - Light&  2.3&  4&  70.4s\\ \hline 
         Average  (ID)&  1.8&  4.8&  66.6s\\ \hline 
          \multicolumn{4}{|l|}{\textbf{State of Object / Location}}\\ \hline 
 Faucet Open& 1.0& 6& 49.5s\\ \hline 
 Stove Open& 1.0& 5& 54.2s\\ \hline 
 Average (State)& 1.0& 5.5& 51.9s\\ \hline 
  \multicolumn{4}{|l|}{\textbf{Surveillance}}\\ \hline 
 Food Burning& 3.1& 4& 62.5s\\ \hline 
 Chemical Spill& 3.7& 5& 55.2s\\ \hline 
 Average (Sur.)& 3.4& 4.5& 58.9s\\ \hline 
  \multicolumn{4}{|c|}{}\\ \hline 
 Average (all)& 2.0& 4.9& 61.6s\\ \hline
        \end{tabular}
    }
    \caption{System performance metrics across benchmarked commands.}
    \label{tab:system_performance_tables}
\end{table}

For the two actuation tasks, we observed a median offset of the drone, from where it was supposed to travel to drop its payload, of 9.1cm and 10.3cm for the ``poison'' and ``medicine'' tasks, respectively. The offset is on orders of centimeters, meaning the system was able to effectively deliver items to the proper location in most cases.

Table~\ref{tab:system_performance_tables} shows statistics about the number of tasks per category that could be performed per full charge of battery, execution time and the number of times the system needed to prompt the user for a more specific description or a more accurate location. We see that the average number of user prompts per command is on average 2. For \textit{state} tasks, this value averaged just 1 prompt (user's initial command). However, for \textit{surveillance} and \textit{ID} tasks, the system often times needed to ask for more information from the user. For all tasks, the average execution time of the visual-language model is on order of seconds, while actuating the drone and analyzing the sensor data is on order of tens of seconds, which is acceptable latency in all of these scenarios. The object/location identification task had a longer average execution time because these tasks generally require the drone to fly and observe multiple locations for each task.%

\subsection{Reconfiguring Mid-Mission}
\label{subsec:deployment_multi}

\begin{figure}[t!]
    \centering
    \includegraphics[width=0.9\linewidth]{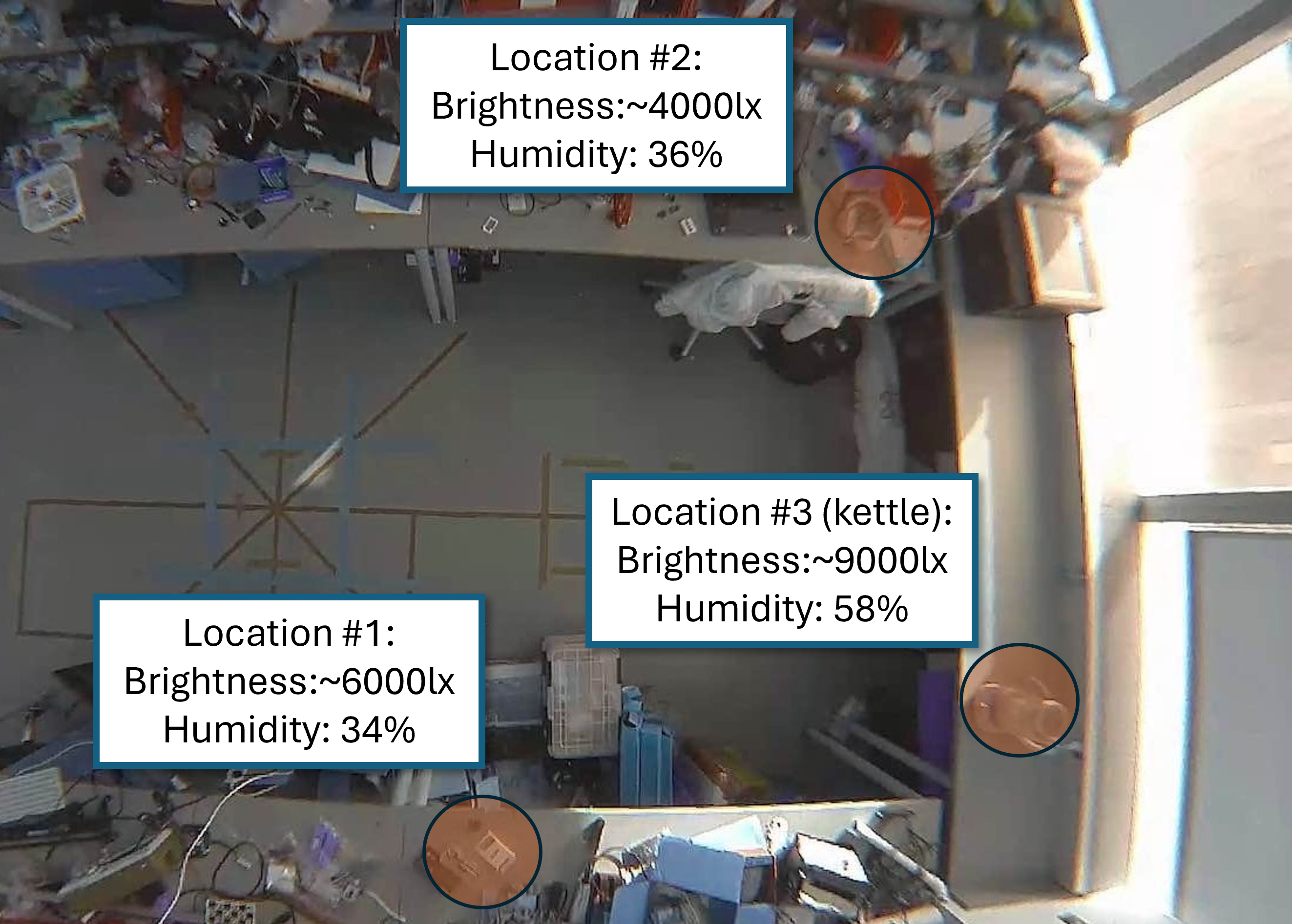}
    \caption{Example of identified locations and measurements in our multi-step ``sense + sense'' task.}
    \label{fig:midmission_story}
\end{figure}

The previous section demonstrated how \newname could be used in conjunction with static sensors in the environment to better perform tasks requiring a single sensor or actuator in a home or office setting. Here, we look at scenarios where a user issues commands that require multiple sensing modalities and actuators, highlighting how \newname could be easily reconfigured mid-mission to satisfy multi-layered tasks. 

Common tasks that require \newname to reconfigure the drone mid-mission come in three different flavors: 1) \textit{actuate + actuate}: multiple individual actuation tasks aggregated into one task (e.g., ``bring me my medicine AND a snack''), 2) \textit{sense + sense}: a task that involves multiple sensing modalities (e.g., ``find me the coolest (temperature) place to sit out of the sunlight (light)'', 3) \textit{sense + actuate}: tasks that involve performing actuation in response to sensing (e.g., ``place rat poison (actuation) in dark areas (sensing)''). Table~\ref{tbl:midmission} summarizes our results, where we show the success rate of the first task (P1), the second task (P2), and the aggregate. We execute each task $20$ times. We discuss each task in more detail, next:

\noindent
\textbf{T1: Actuate + Actuate - ``Bring medicine AND snack''}. In this task, the user wants two (P1 and P2) items brought to him/her. We see that the success rate of delivering both items is high, just like we observed in Section~\ref{subsec:deployment_lab}. The second item (snack) failed two times (P2) because the swapping failed; the drone landed with a high offset from the center of the landing station (Figure~\ref{fig:takeoff_landing}d). 

\noindent
\textbf{T2: Sense + Actuate - ``Put poison in the warmest area''}. In this task, the user may want to place poison (P2) for rats and bugs in warm areas (P1) where they are likely to congregate (e.g., warm places). We simulate ``warm places'' boiling water in a kettle in locations visible to cameras. There were three times that a location away from the kettle was chosen (P1). In these instances, our image segmentation approach cut out the area we placed the kettle, so the failure point was from the camera rather than \newname. Improving sensing with foundational models in future work is key to realizing a robust version of this end-to-end system.  

\noindent
\textbf{T3: Sense + Sense - ``What is the most humid and brightest location for placing a plant''}. Users may want to know a humid (P1) and bright (P2) place for optimal plant growth. We simulate the ``correct'' location by placing a lamp and kettle at the desired location. There was one instance where the task needed human intervention (failed) because of unsuccessfully swapping in the second sensor; again, the failure came from landing the drone (P2). The foundational model correctly identified the location we placed the lamp, and the drone flew to these locations with the humidity sensor. However, we stopped the run when the drone landed with a high offset that the funnel-shaped landing station could not realign. Another instance failed because our foundational model pipeline did not identify the location we simulated high brightness and humidity (P1). Figure~\ref{fig:midmission_story} shows an example of a successful run, displaying points where the foundational model identified to send the drone, as well as humidity and light measurements  taken by the drone for each of these locations to make the final location determination. Here, we placed the kettle at location three.

\begin{table}[!t]
	
	\centering
	\resizebox{.45\textwidth}{!}{
		\begin{tabular}{|l|c|c|c|c|c|}
			\hline
		\textbf{Category}	& \textbf{Success P1} & \textbf{Success P2} & \textbf{Success Total}\\
			
			\hline
			\textbf{T1: Actuate + Actuate} & 20/20 = 1.0 & 18/20 = 0.90 & 18/20 = 0.90\\
			
			\textbf{T2: Sense + Actuate} & 17/20 = 0.85 & 20/20 = 1.0 & 22/25 = 0.85\\
			\textbf{T3: Sense + Sense} & 19/20 = 0.95 & 18/19 = 0.95 & 18/20 = 0.90\\

			\hline
		\end{tabular}
	}
	 
      \caption{Summary of success rate for tasks that require reconfiguring the drone mid-mission, broken down by success rate of the first leg (P1) and second leg (P2).}
	\label{tbl:midmission}
\end{table}

\begin{table}[!t]
	
	\centering
	\resizebox{.42\textwidth}{!}{
		\begin{tabular}{|l|c|c|c|c|c|}
			\hline
		\textbf{Category}	& \textbf{\# Prompts} & \textbf{\# Exec} & \textbf{Exec Time (s)} & \textbf{Success}\\
  
        	& & & \textbf{(vLLM + drone)} & \\
			
			\hline
			\textbf{ID} & 1.0 & 21 & 0.51 + 51.30 & 15/21 = 0.71\\
			
			\textbf{State} & 1.0 & 13 & 0.13 + 37.20 & 11/13 = 0.85\\
			\textbf{Surveil.} & 1.2 & 22 & 0.21 + 43.87 & 19/22 = 0.87\\
   
            \textbf{Actuation} & 2.1 & 31 & 0.27 + 29.50 & 31/31 = 1.0 \\
			
			\hline
		\end{tabular}
	}
	 
  \caption{Summary of tasks during in-the-wild deployments. The number of prompts is the average number of times the user needed to prompt the system. This number is often greater than one because either the user used a non-specific adjective (e.g., ``best'' rather than ``warmest'' location) or the system needed to narrow down the potential candidate locations in the case of actuation tasks. Number of executions is the total number of tasks issued during the deployment period.}
	\label{tbl:inthewild_deployment}

\end{table}

\subsection{In-the-Wild Deployment}
\label{subsec:deployment_wild}

After benchmarking several tasks per category, we allowed people who occupied this office space (Figure~\ref{fig:localization_benchmark}e) to freely use the system over the course of $5$ days. Table~\ref{tbl:inthewild_deployment} summarizes the number of events that occurred during this period. A total of 8 people issued 87 commands to the system during this time period.

We see that most of the actions issued throughout the deployment were actuation tasks. Around 90\% of these tasks involved bringing the user a snack, which we loaded and manually refilled into actuation modules throughout the deployment. \textit{ID} tasks that users issued generally fell into two categories: finding an area with the least amount of sunlight (our space has many windows and is susceptible to glare) or finding a lost item (e.g., a wallet or phone). For the \textit{surveillance} and the \textit{state} tasks, most users asked the system about a 3D print job, whether a heat element was left on (e.g., soldering iron), or if there were anyone occupying different parts of the space. 

The category of tasks that had the lowest success rate was the object/location ID category. This is because most of these tasks relied on static cameras or the camera on the drone to find something extremely small in the landscape of a scene (e.g., a circuit component or a phone), making it difficult for the visual-language pipeline to identify relevant locations. Even after flying the drone to the specified location, it can be difficult to detect; we envision future work focusing on how to design search algorithms and protocols for drones to identify small objects of interest. Several items that users wanted the drone to look for were also underneath furniture or tables; a camera mounted on the ceiling or walls have limited view of these items. Another avenue of research for realizing a drone or robot-based personal assistant could be how to leverage and design small robotic systems (e.g., physical design, path planning, search algorithms, etc.) to reach and look for items in areas unobservable by static sensor deployments. On the flip side, the actuation task had the highest success rate particularly because the system prompted users each time to confirm the location to make the delivery, which reduces reliance on language models and perception algorithms to make this determination. Although there are still improvements needed to realize a truly autonomous drone-based personal assistant, all users were positively receptive to this system and could see its value.

\section{\add{Discussion and Future Work} }
\label{sec:discussion}
\add{
\paragraph{Usability}The deployment of autonomous drones in indoor environments raises critical usability challenges. A primary concern is noise disruption from drone propellers, which could be mitigated through: 1) flight path optimization that maintains higher operating altitudes when possible, reducing perceived ground-level noise~\cite{adlakha2023integration}; 2) implementation of low-noise propeller designs that recent aerodynamics research suggests could reduce noise by around 5dB~\cite{sebastian2020toroidal,jansen2024impact}; and 3) context-aware navigation that avoids occupied areas during noise-sensitive periods. Beyond noise, user interaction with the system requires streamlining. We envision developing natural interaction paradigms including gesture control for intuitive drone guidance~\cite{abtahi2017drone}, and an augmented reality interface for visualizing drone intentions and planned paths~\cite{mourtzis2022unmanned,konstantoudakis2022drone}.
\paragraph{Privacy}
Privacy is a critical concern for camera networks and camera-equipped drones in indoor spaces. In this work, camera feeds are transmitted locally and processed on a local server for scene understanding, vision-language grounding and drone navigation. The need for server can be bypassed as more efficient, compact foundation models and powerful edge computers emerge, as well as leveraging compression techniques such as quantization and distillation to create dedicated smaller models~\cite{sharshar2025edgeVLMSurvey, cai2024VLM} for each vision task mentioned above while preserving the generalizibility.
\paragraph{Extending to Diverse Environments}
While we We demonstrated the effectiveness of our implementation in several indoor environments, extending FlexiFly to new settings presents unique challenges and opportunities. Flexifly can be adapted to environments with existing camera infrastructure as long as the relative position of the cameras are known; this can be achieved in a self-supervised manner by leveraging recent advances in camera self-localization and calibration techniques~\cite{liu2022sofit}, potentially enhanced by fine-tuning vision-language models for specific deployment contexts. However, we acknowledge significant limitations in environments where camera deployment is impractical or restricted. For spaces primarily monitored by non-visual sensors (e.g., RF, vibration, or acoustic sensors), new methodologies beyond ARCK-Means and our vision pipeline must be developed for localizing areas of interest and decision-making. We envision that future FMs and penetrative AI~\cite{xu2024penetrative} could help process diverse sensor modalities spread throughout environments, aided by novel segmentation approaches for non-visual spatial data to enable accurate drone navigation.
}

\section{Conclusion \delete{and Future Work}}\label{sec: conclusion}
Our work studies the possibility of LLMs and FMs as a general intelligence for physical spaces. We identify that FMs analyzing sensing data monitoring a large space have difficulty identifying localized events that occur in small areas. As such, we propose novel segmentation methods and drones with reconfigurable sensing and actuation that enable FMs to identify and ``zoom in'' to analyze targeted areas with higher resolution. We demonstrate through a real deployment of a personal assistant application that \newname can improve the successful completion of complex tasks throughout our physical spaces by up to $85\%$. \delete{While this work focuses on leveraging camera networks and VLMs, due to their maturity, we envision that future FMs and penetrative AI~\cite{xu2024penetrative} could process modalities from other diverse sensors spread throughout environments, as they become more mature. Moreover, we anticipate similar \textbf{advancements in indoor navigation for drones} could enable general intelligence to ``zoom in'' with greater speed, efficiency, and accuracy. We leave these aspects to future work. }\newname is a critical step towards FMs and LLMs that can naturally interact and actuate the physical environment, just as they have shown in many applications in the digital domain.

\begin{acks}
This research was partially supported by COGNISENSE, one of seven centers in JUMP 2.0, a Semiconductor Research Corporation (SRC) program sponsored by DARPA, as well as the National Science Foundation under Grant Number CNS-1943396. The views and conclusions contained here are those of the authors and should not be interpreted as necessarily representing the official policies or endorsements, either expressed or implied, of Columbia University, NSF, SRC, DARPA, or the U.S. Government or any of its agencies.
\end{acks}

\balance
\bibliographystyle{ACM-Reference-Format}
\bibliography{references}

\end{document}